%% file: iclr2027_conference.tex
\documentclass{article}
\usepackage{iclr2027_conference,times}

\usepackage{amsmath,amsfonts,bm}

\usepackage[hidelinks]{hyperref}
\usepackage{url}
\usepackage{graphicx}
\usepackage{cleveref}
\usepackage{xcolor}
\usepackage{booktabs}

\title{Look, Reason, Draw a Region: Recursive \mbox{Refinement} for Spatial Context Prediction}

\author{%
Wajeeha Tahir$^{1}$ \And
Anna Tegon$^{1}$ \And
Lionnus Kesting$^{1}$ \And
Gamze İslamoğlu$^{1}$ \AND
\makebox[\dimexpr\textwidth-2\tabcolsep][c]{%
    Kristóf Karacs$^{2}$\qquad\qquad
    Luca Benini$^{1}$\qquad\qquad
    Thorir Mar Ingolfsson$^{1}$%
}\\[5pt]
\makebox[\dimexpr\textwidth-2\tabcolsep][c]{%
\normalfont $^{1}$Integrated Systems Laboratory, ETH Z{\"u}rich, Switzerland}\\
\makebox[\dimexpr\textwidth-2\tabcolsep][c]{%
\normalfont $^{2}$Pázmány Péter Catholic University, Hungary}
}

\iclrfinalcopy
\renewcommand{\headrulewidth}{0pt}
\begin{document}

\maketitle

\input{text/00_abstract.tex}

\input{text/01_intro.tex}
\input{text/02_background.tex}
\input{text/03_problem.tex}
\input{text/04_arch.tex}
\input{text/05_evaluation.tex}

\input{text/06_benchmarking.tex}

\input{text/08_discussion.tex}

\subsection*{AI use statement}
In this work, we used generative AI tools, including Anthropic Claude and OpenAI Codex, to assist with developing utility scripts, and code refactoring. We also used generative AI tools for language editing to improve the clarity and readability of author-written text. We also used these tools to search for and summarize related work, which helped us identify papers to read. AI tools were not used to generate data, formulate research questions or experimental design, develop models, interpret results, conduct literature review, or generate figures. All AI-assisted code and text changes were reviewed and validated by the authors. The authors take responsibility for the final content of this work, including text, code and scientific claims.

\subsection*{Reproducibility statement}
For dataset generation, we build on the RoboSpatial~\citep{song2025robospatial} codebase. The generation procedure, including modifications introduced to support our task, is detailed in~\Cref{app:training-data}. For the model architecture and training pipeline, we extend the Tiny Recursive Models~\citep{jolicoeur2025less} codebase. Architectural components and recursive schedules are detailed in~\Cref{app:arch}; checkpoint selection and training configuration are described in~\Cref{app:checkpoint-selection,app:training-setup}.

\subsubsection*{Acknowledgments}
This work was supported by the Hasler Foundation under the Small Project “Quantized Recursive Inference for Resource-Constrained Edge AI” (project 2026-03-25-875), and by a grant from the Swiss National Supercomputing Centre (CSCS) under project ID lp160 on Alps.

\begingroup
\small
\bibliography{iclr2027_conference}
\bibliographystyle{text/iclr2027_conference}
\endgroup

\input{text/appendix.tex}

\end{document}

%% file: text/00_abstract.tex
\begin{abstract}
    Spatial context prediction requires visual grounding and spatial reasoning to identify vacant regions relative to objects in a scene. Spatial relationships and physical constraints remain challenging for large vision-language models (VLMs). Recursive reasoning has demonstrated strong performance on discrete and abstract reasoning tasks, motivating its use for improving spatial context prediction in natural images. Weight sharing enables higher accuracy with a small trainable parameter count. We introduce Spatial-TRM, a compact recursive model with frozen vision-language features, a convex region decoder, and just 11.2M trainable parameters. This weight-tied reasoner supports six spatial relationships, three reference frames, and open-vocabulary anchor objects. On the RoboSpatial-Home benchmark, Spatial-TRM achieves 29.51\% context accuracy, a gain of 9.02 percentage points (44.0\% relative) over a separately trained single-step model with the same parameter count. Including the frozen encoders, the model has 877.8M total parameters, approximately $15\times$ fewer than the reported 13B size of RoboPoint. These results show that repeated refinement improves spatial context prediction from pretrained vision-language features at a fixed trainable parameter count on this benchmark.
\end{abstract}

%% file: text/01_intro.tex
\section{Introduction}
An instruction such as ``find free space to the left of the bowl'' requires more than locating the bowl~\citep{yuan2025robopoint}. A model must identify the anchor object, interpret the requested direction in the appropriate reference frame, and distinguish vacant space from occupied regions~\citep{song2025robospatial}. We study spatial context prediction: given an RGB image and a spatial prompt, predict a candidate region that satisfies the requested relationship. We ask: can repeated refinement improve spatial-context prediction from frozen features at a fixed parameter count?

Recursive refinement revisits a prediction using a compact, shared network that iteratively updates its internal representations~\citep{jolicoeur2025less}. Instead of adding parameters for each processing step, these architectures scale reasoning depth by repeatedly applying the same weights~\citep{dehghani2019universal}. Tiny Recursive Models (TRMs) have demonstrated this approach on discrete reasoning tasks~\citep{jolicoeur2025less}. We investigate this form of refinement for spatial context prediction in natural images.

In our setting, these updates must connect object grounding with directional and geometric constraints~\citep{song2025robospatial}. Pretrained vision-language encoders provide aligned visual and semantic features~\citep{jose2025dinov2,simeoni2025dinov3}. Existing spatial models learn task-specific outputs through spatial question answering~\citep{chen2024spatialvlm} or instruction-conditioned pointing~\citep{yuan2025robopoint}.  

In this work, we introduce Spatial-TRM, a compact model that combines latent recursive refinement with explicit region prediction. Given frozen, open-vocabulary image and text features~\citep{simeoni2025dinov3,jose2025dinov2}, a small weight-tied model grounds the anchor object and refines a latent region representation through repeated updates using 11.2M trainable parameters. A geometric decoder maps the refined representation to a convex mask by intersecting predicted half-planes~\citep{deng2020cvxnet}. The model supports six spatial relationships and three reference frames. 

We train on the spatial context subset of RoboSpatial~\citep{song2025robospatial} and evaluate transfer to RoboSpatial-Home. On its 122 context questions, Spatial-TRM achieves 29.51\% context accuracy with 877.8M total parameters, approximately $15\times$ fewer than the reported 13B size of RoboPoint~\citep{yuan2025robopoint,song2025robospatial}. We also examine how representative-point selection and incomplete target masks affect the interpretation of region predictions.

Our contributions and main findings are:
\begin{itemize}
\item We adapt latent recursive refinement to spatial context prediction, improving context accuracy from 20.49\% to 29.51\% over a separately trained single-step model with the same parameter count and frozen image and text features. This is a 44.0\% relative increase in accuracy without adding parameters.
\item We build a region predictor with fewer than one billion total parameters by combining frozen image and text encoders, recursive refinement, and a convex decoder. With a CLIP backbone~\citep{radford2021clip}, the full model uses 438M parameters and achieves 26.23\% accuracy: half the size of the DINOv3 variant at a cost of 3.28 percentage points.
\item We evaluate transfer to RoboSpatial-Home and quantify sensitivity to representative-point selection: increasing the point budget from one to ten raises accuracy from 29.51\% to 44.26\% for the same predicted masks. The key distinction is between agreement with the annotations and validity of the entire predicted region; point accuracy and mask overlap do not establish the latter on their own.
\end{itemize}

%% file: text/02_background.tex
\section{Related Work}

\paragraph{Spatial reasoning and referring.}
SpatialVLM learns qualitative and quantitative spatial relations through spatial question answering~\citep{chen2024spatialvlm}. RoboPoint predicts image-space points for language-specified spatial affordances, including vacant locations relative to objects~\citep{yuan2025robopoint}. RoboSpatial evaluates configuration, context, and compatibility across different reference frames~\citep{song2025robospatial}. RoboRefer introduces a dedicated depth encoder and combines supervised and reinforcement fine-tuning for spatial referring~\citep{zhou2025roborefer}, while RoboBrain 2.0 combines spatial perception with broader embodied reasoning and planning~\citep{baai2025robobrain}. Our focus is narrower: learning a region predictor for spatial-context queries from frozen RGB and text features.

\paragraph{Frozen visual and language representations.}
DINOv2 Meets Text aligns visual features with text at both image and pixel levels~\citep{jose2025dinov2}. DINOv3 provides transferable visual features and extends this text-alignment approach to its representations~\citep{simeoni2025dinov3}. These models provide the features used by Spatial-TRM; the grounding, recursive refinement, and region prediction modules are trained for the spatial task. Freezing the encoders limits the number of task-specific trainable parameters, while retaining the cost of the pretrained encoders in the complete system.

\paragraph{Recursive refinement.}
Universal Transformers reuse a transformation across processing steps, tying parameters across depth~\citep{dehghani2019universal}. Tiny Recursive Models use a compact shared network to update latent reasoning and answer states, with evaluations on Sudoku, maze solving, and abstract reasoning tasks~\citep{jolicoeur2025less}. We adapt this latent refinement mechanism to natural-image spatial-context prediction. Weight sharing permits additional refinement without a separate parameter set for each step; its effect on prediction quality and computational cost must be evaluated separately.

\paragraph{Convex region representations.}
CvxNet represents three-dimensional shapes through learnable convex components defined by half-space constraints~\citep{deng2020cvxnet}. Our decoder uses a related construction in two dimensions to produce a single convex region. The contribution is the combination of this geometric representation with recursive refinement for spatial-context prediction. 

%% file: text/03_problem.tex
\section{Problem Formulation}

\begin{figure}[t]
  \centering
  \includegraphics[width=\linewidth]{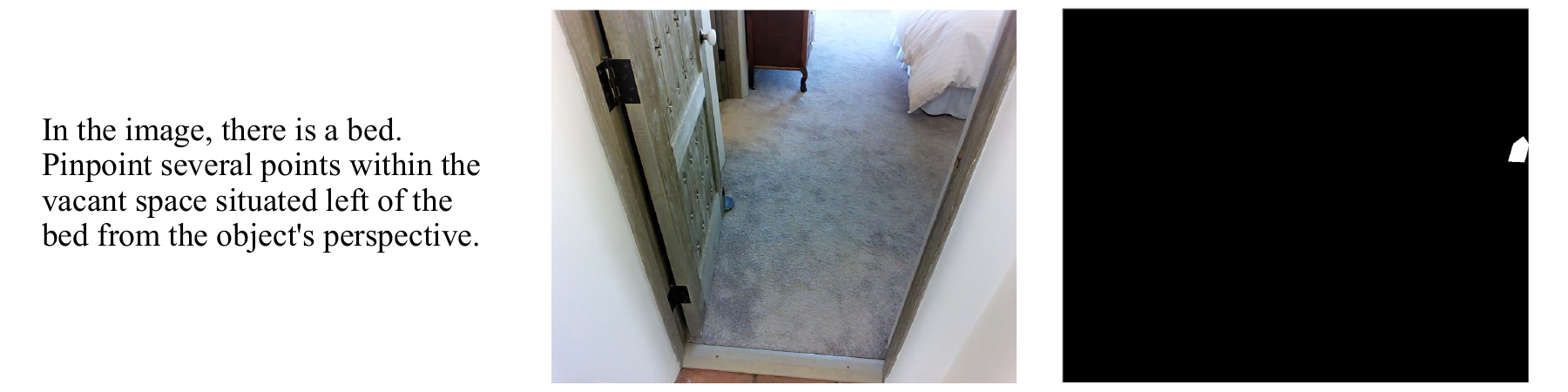}
  \caption{Example of a spatial-context query and its region target. The prompt specifies the bed as the anchor and asks for vacant space to its left in the object's reference frame. The mask is constructed from the annotated target points.}
  \label{fig:data_sample}
\end{figure}

The RoboSpatial benchmark formulates spatial context prediction as a pointing task, where the expected output is a set of coordinates identifying valid locations in the scene~\citep{song2025robospatial}. We formulate the task as binary segmentation, with a mask describing the candidate region. This representation allows the model to refine a candidate region using spatial features from the entire image. We focus on context queries that specify a region relative to an anchor object and a reference frame~\citep{song2025robospatial}. A mask describes both the location and extent of a candidate area. \Cref{fig:data_sample} illustrates the problem formulation and the expected output.

To construct region targets, we enclose the annotated RoboSpatial target points~\citep{song2025robospatial} in a two-dimensional convex hull. The resulting mask need not cover all valid parts of the spatial region. Conversely, a convex hull can also span occupied areas between annotated points. These masks therefore provide supervision derived from sparse annotations, rather than complete labels of vacant space.

%% file: text/04_arch.tex
\section{Spatial-TRM}

Spatial-TRM first grounds the anchor, then repeatedly updates a latent spatial prediction, and finally decodes a candidate region. The architecture combines a frozen encoder backbone, a cross-attention grounding step, a recursive reasoning block with a halting head used during training, and a final decoder that reconstructs a convex region from the refined latents.
The vision and text encoders account for 866.6M parameters which we do not fine-tune. The cross-attention grounding module contains 2.1M, while the core reasoning block and decoder contribute 6.8M and 2.3M parameters, respectively. In total, Spatial-TRM contains 877.8M parameters of which only 11.2M are trainable, equivalent to 1.3\% of the total model size. 

\subsection{Input Representation}
We use the DINOv3 ViT-L/16 encoder with the dino.txt text tower for its fine-grained patch-level features and open-vocabulary object localization~\citep{simeoni2025dinov3, jose2025dinov2}. The encoder weights remain frozen during training. For each input prompt, we extract two distinct text embeddings: the first, utilized for object grounding, is derived solely from the anchor object name, and the second full-prompt embedding provides the semantic context required for spatial reasoning.

RoboSpatial context questions follow standardized templates~\citep{song2025robospatial}, which reduces variation in how spatial queries are phrased. For these prompts, the dino.txt encoder provides text embeddings for both grounding and reasoning while keeping its parameters fixed.

For object grounding, using a multi-head cross-attention layer, the image patch tokens query the anchor object text tokens. The resulting anchor-conditioned patch features are added residually to the initial image representations. Finally, these refined features are then concatenated with the full-prompt embedding to form the input latent state for the subsequent reasoning block.

\subsection{Latent Recursive Refinement}
We build our recursive reasoning core on the latent refinement mechanism introduced by the Tiny Recursive Model (TRM)~\citep{jolicoeur2025less}, adapting it from discrete problems to visual spatial reasoning. The reasoner maintains a reasoning state $z$ and an answer state $y$, each with the same dimensionality as the input representation. The input latent is repeatedly injected into $z$, which is then used to update $y$. Both state updates apply the same two-layer transformer, so successive refinements reuse the same parameters. We apply two-dimensional rotary positional encodings~\citep{su2024roformer} to the vision tokens to preserve spatial information, and one-dimensional encodings to the text tokens. Self-attention processes the joint image and text sequence at each update.

Within each deep recursion cycle, four updates of $z$ precede one update of $y$. We execute three such cycles per deep-supervision step and repeat deep supervision up to four times during training. The answer state supplies the refined representation for region decoding. Training and gradient-flow details are given in~\Cref{app:reasoning}. 

To determine when to stop refinement during training, we follow TRM's learned halting approach~\citep{jolicoeur2025less}, inspired by adaptive computation time (ACT)~\citep{graves2017adaptivecomputationtimerecurrent}. The head is implemented as a linear layer that predicts a halting logit from the current answer latent at each deep-supervision step; the logit is then used to determine whether to halt early rather than executing all four steps. The halt mechanism is trained using an auxiliary loss, which is weighted and combined with the primary task objective. During inference, this adaptive halting is disabled and the model utilizes the full deep-supervision budget. 

\subsection{From Latents to Masks}
\begin{figure}[t]
  \centering
  \includegraphics[width=0.65\linewidth]{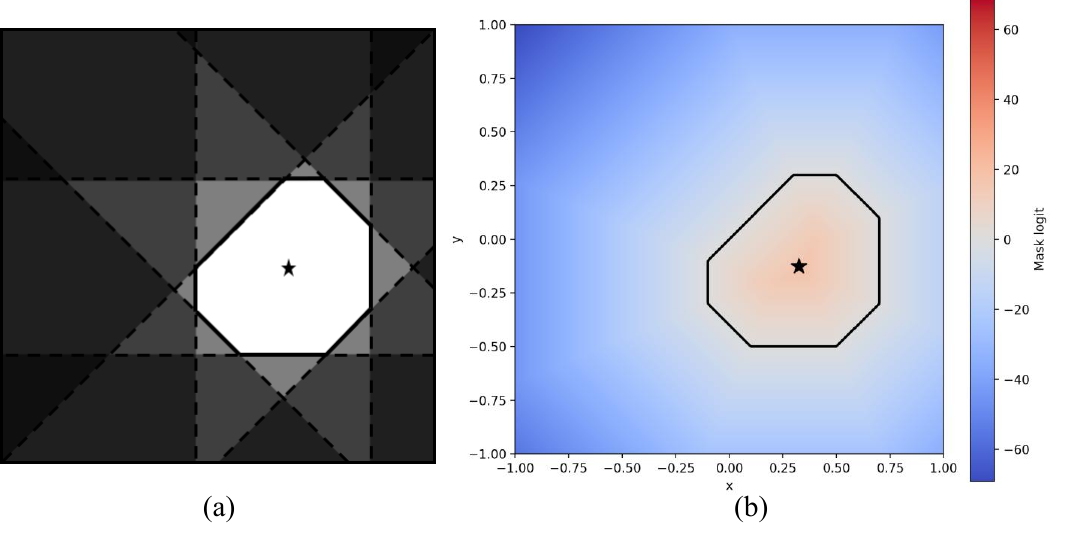}
  \caption{(a) The decoder predicts an interior point and distances to half-plane boundaries at fixed normal directions, whose intersection defines a convex polygon. (b) Each pixel is assigned a logit based on its minimum margin across the half-planes.}
  \label{fig:convex-decoder}
\end{figure}

Inspired by the half-space representation used in CvxNet~\citep{deng2020cvxnet}, the decoder predicts an interior reference point and a set of half-plane offsets at fixed, evenly spaced normal directions. Their intersection defines a single convex polygon representing the predicted foreground region. This parameterization guarantees that the predicted region is connected and does not contain holes by construction. With $K=32$, the decoder predicts 34 geometric values (two interior-point coordinates and 32 offsets) that define a $224\times224$ mask. 

The decoder context combines visual features from the encoder, refined latent representations from the reasoner, and normalized patch coordinates. These context tokens provide information about scene content, inferred spatial relationships, and the location of each patch. The decoder uses K+1 learned query tokens: one to predict the interior point while the remaining K queries predict half-plane distances. We assign the k-th boundary a fixed normal direction 
\[
n_k = (\cos\theta_k, \sin\theta_k), \quad \theta_k = 2\pi k/K
\]
such that the normals are evenly spaced. 

This normal encoding is added to each half-plane query to assign distinct geometric roles.  

The interior point and distance queries first interact through self-attention and then cross-attend to the context tokens. The predicted reference point is not necessarily the geometric centroid of the polygon, but rather a point guaranteed to lie inside the foreground region. Each directional query predicts the perpendicular distance from the reference point to its corresponding half-plane boundary. 

For every pixel p, a signed margin is computed with respect to each half-plane boundary:
\[
m_k(p) = d_k - n_k (p - c)
\]
where $c$ is the interior point, $n_k$ is the normal direction of the k-th half-plane, and $d_k$ is the boundary offset. A positive margin indicates the pixel satisfies the corresponding half-plane constraint. Because a foreground pixel must satisfy all half-plane constraints, the minimum margin $\min_{k=1,\ldots,K} m_k(p)$ across all boundaries is used to identify foreground pixels. These minimum margin logits, when thresholded at 0, produce binary masks which are convex by construction.

\subsection{Training Objectives}
Mask fidelity is optimized using a class-weighted binary cross-entropy loss between the predicted and ground truth masks. Foreground pixels are assigned a higher weight to address the class imbalance due to valid regions being a minority in the image. 

The halting head is trained with a separate binary cross-entropy loss to predict whether the current prediction satisfies the spatial agreement criterion described below. 

We also use a relational contrastive loss to encourage the model to differentiate between spatial relations through prompt-conditioned predictions. For each training sample, we generate a negative prompt by replacing the spatial relation with a randomly picked one (e.g., "left" may become "above"). We require the correct relation prediction to exhibit more spatial agreement with the ground truth compared to the negative relation prediction. 

For this comparison, we average pool the predicted and ground-truth masks to a coarser grid and compare their flattened representations using cosine similarity $\text{sim}(p,t)$ between prediction $p$ and target $t$. Pooling details are given in~\Cref{app:objectives}.

With margin $m$, the relational contrastive loss is:
\[
\mathcal{L}_{\mathrm{rel}} = \max(0, m - \text{sim}(p_{correct}, t) + \text{sim}(p_{negative}, t))
\]

As the annotations do not provide a ground truth for the negative spatial relation, this loss only enforces relative alignment with the target relation without providing a valid-region target for the negative prompt.

\subsubsection{Adaptive Halting Target}
As the deep-supervision cycles in the reasoning block can run for the full reasoning budget, early halting can reduce the number of updates executed during training. For this purpose, the ACT head requires a correctness criterion to determine if the current answer is sufficiently valid for the recursion to halt. 

We experimented with two extremes for defining this criterion: (1) whether the predicted centroid falls within the ground truth mask and (2) whether the prediction achieves an IoU greater than 0.5. The former is permissive because a single point does not constrain the full spatial extent. Conversely, the latter is overly restrictive, and since only a few predictions achieve sufficient overlap, the model consumes nearly the entire recursion budget. 

We use an intermediate criterion that complements the centroid-based target with additional support points, sampled as the 8-connected neighbors of the centroid at a fixed distance. Rather than requiring exact overlap, this metric measures mutual spatial agreement. We compute the criterion bidirectionally: from the prediction to ground truth and vice versa, as a proxy for both spatial relationship and contextual understanding.

%% file: text/05_evaluation.tex
\section{Evaluation Protocol}
Conceptually, a predicted mask is considered valid if it satisfies two conditions:

\begin{enumerate}
   \item it correctly identifies the perspective and direction relative to the anchor object, and
   \item the identified vacant space does not intersect with any objects and is in the general vicinity of the object.
\end{enumerate}

\paragraph{Context accuracy}
We measure context accuracy using the point-in-mask criterion of RoboSpatial-Home~\citep{song2025robospatial} by verifying whether the decoder's single predicted interior point lies within the ground truth mask. A point-based match does not establish that the full predicted region satisfies the requested relationship or is unoccupied. The score also depends on representative-point selection as discussed in~\Cref{sec:ood_benchmarks}.

\paragraph{Mask overlap}
We also report intersection over union (IoU) and Dice coefficient to quantify overlap with the annotated target. The prompts require the model to identify a valid vacant region, for which the ground-truth mask may cover only part of the valid space. 

A prediction that satisfies the intended spatial relationship may receive a low score if its extent and boundaries do not align with the annotated target. Low metric scores can reflect either annotation mismatch or prediction error. Examples and implications for training are provided in~\Cref{app:evaluation}.

%% file: text/06_benchmarking.tex
\section{Out-of-Distribution Generalization}
\label{sec:ood_benchmarks}
We evaluate Spatial-TRM on the RoboSpatial-Home benchmark which uses a similar prompt template but omits perspective information, introducing ambiguity in the reference frames. Its scenes are also more cluttered compared to the training set, which contains approximately three objects per image on average, thus challenging both open-vocabulary object grounding and spatial reasoning. Of the benchmark questions, 122 correspond to the spatial context tasks that this work covers, on which we report context accuracy. Our main result uses 40,000 training samples, the centroid-based checkpoint-selection heuristic in~\Cref{app:checkpoint-selection}, and one predicted interior point for benchmark scoring. The separate image-disjoint training control is reported in~\Cref{app:training-data}. Benchmarks requiring zero-shot generalization on relations not present in the training vocabulary are outside the scope of our evaluation.

\paragraph{Recursion improves accuracy at a fixed parameter count}
To isolate the effect of repeated refinement, we train a separate model with the reasoning core retained and recursion disabled: the model executes a single reasoning step followed by a single answer state update. The architecture is otherwise unchanged, and the model has the same number of parameters as the recursive model. In~\Cref{tab:recursion-ablation}, this variant achieves 20.49\% context accuracy, compared with 29.51\% for the recursive model: a gain of 9.02 percentage points, or 44.0\% relative.

\begin{table}[htbp]
\caption{Reasoning ablation on RoboSpatial-Home. The separately trained single-step and recursive models both have 11.2M trainable parameters; repeated refinement improves context accuracy by 9.02 percentage points. All variants use the same frozen encoder.}
\label{tab:recursion-ablation}
\centering
\small
\setlength{\tabcolsep}{7pt}
\renewcommand{\arraystretch}{1.12}
\begin{tabular}{@{}lrr@{}}
\toprule
\textbf{Model} & \textbf{Trainable parameters} & \textbf{Accuracy (\%)} \\
\midrule
Feedforward & 4.46M & 17.21 \\
Single-step & 11.2M & 20.49 \\
\textbf{Recursive (ours)} & 11.2M & \textbf{29.51} \\
\bottomrule
\end{tabular}
\end{table} 

\begin{figure}[t]
  \centering
  \includegraphics[width=\linewidth]{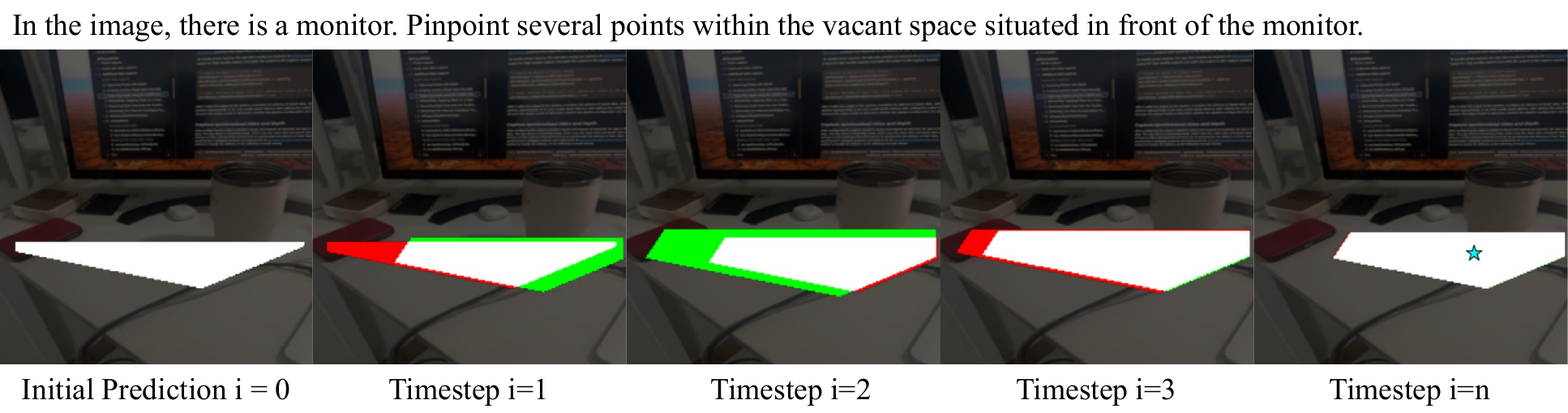}
  \caption{Through iterative updates, the recursive model updates its current answer state over a fixed budget. Foreground regions added at each step are represented in green and removed regions in red. The final prediction at step $n$ is evaluated using the interior point marked by a star.}
  \label{fig:mask-progression}
\end{figure}

This comparison holds parameter count fixed, but the recursive model uses additional computation. Changes in the predicted region are illustrated in~\Cref{fig:mask-progression}. 

As an additional control, we compare Spatial-TRM against a feedforward network with the same frozen encoder. The image patches attend over the full prompt, the result of which is added to the image features. We bypass the recursive reasoning core entirely and construct the decoder context from the refined image features. With 2.1M parameters for the fusion layer and 2.36M parameters for the decoder, the feedforward model has a total of 4.46M trainable parameters. Trained with the same configuration, dataset and task losses, the feedforward model scores 17.21\% context accuracy compared with 29.51\% for the recursive model, corresponding to a difference of 12.30 percentage points. This feedforward comparison also changes the trainable parameter count. 

\paragraph{Comparison with larger models}
In~\Cref{tab:robospatial-home}, we compare Spatial-TRM against pointing models, embodied spatial reasoning models, proprietary foundation models and open-weight VLMs, a subset of which are fine-tuned on RoboSpatial. Baseline scores are taken from the sources specified below the table. They were not re-evaluated under a common protocol; prompting, point selection, and model versions may differ across sources.

\begin{table}[h]
\caption{RoboSpatial-Home context accuracy (higher is better). Baseline parameter counts are nominal model sizes; the SpaceTools count excludes external tools. Our count includes the frozen encoders. $^{*}$ denotes fine-tuning on RoboSpatial; a dash indicates an unreported parameter count.}
\label{tab:robospatial-home}
\centering
\small
\setlength{\tabcolsep}{6pt}
\renewcommand{\arraystretch}{1.08}
\begin{tabular}{@{}lrr@{}}
\toprule
\textbf{Model} & \textbf{Accuracy (\%)} & \textbf{Parameters} \\
\midrule
Qwen2.5-VL-3B~\citep{bai2025qwen25vltechnicalreport} & 0.00 & 3B \\
Molmo-7B~\citep{Deitke_2025_CVPR} & 0.82 & 7B \\
Qwen2.5-VL-32B~\citep{bai2025qwen25vltechnicalreport} & 3.28 & 32B \\
GPT-4o~\citep{openai2024gpt4ocard} & 6.60 & - \\
Gemini 2.5 Flash~\citep{comanici2025gemini25pushingfrontier} & 7.90 & - \\
Gemini 2.5 Pro~\citep{comanici2025gemini25pushingfrontier} & 8.30 & - \\
GPT-5-mini~\citep{singh2026openaigpt5card} & 12.50 & - \\
SpaceLLaVA-13B~\citep{chen2024spatialvlm}$^{*}$ & 13.10 & 13B \\
VILA-1.5-8B~\citep{Lin_2024_CVPR}$^{*}$ & 15.60 & 8B \\
GPT-5~\citep{singh2026openaigpt5card} & 19.00 & - \\
LLaVA-NeXT-8B~\citep{li2024llavanextstrong}$^{*}$ & 19.70 & 8B \\
Claude Sonnet 4.5~\citep{anthropic2025sonnet45} & 23.77 & - \\
\midrule
\textbf{Spatial-TRM (ours)}$^{*}$ & \textbf{29.51} & \textbf{877.8M} \\
\midrule
RoboPoint-13B~\citep{yuan2025robopoint}$^{*}$ & 31.10 & 13B \\
Gemini-ER 1.5-thinking~\citep{geminiroboticsteam2025geminirobotics15pushing} & 31.10 & - \\
RoboBrain2.0-7B~\citep{baai2025robobrain} & 44.35 & 7B \\
SpaceTools-3B~\citep{chen2026spacetools} & 52.46 & 3B \\
RoboRefer-8B-SFT~\citep{zhou2025roborefer} & 61.48 & 8B \\
\bottomrule
\end{tabular}
\par\vspace{3pt}
{\footnotesize Spatial-TRM: 866.6M frozen + 11.2M trainable parameters. Scores for the four fine-tuned baselines: RoboSpatial, Table~4 (arXiv v5)~\citep{song2025robospatial}; Gemini and GPT-5 models: Gemini Robotics 1.5, Table~19~\citep{geminiroboticsteam2025geminirobotics15pushing}; GPT-4o: the RoboSpatial-Eval leaderboard~\citep{song2026robospatialeval}; remaining baselines: SpaceTools, Table~2~\citep{chen2026spacetools}. SpaceLLaVA is a community implementation of SpatialVLM~\citep{song2025robospatial}.\par}
\end{table}

Spatial-TRM achieves a context accuracy of 29.51\%, exceeding the reported scores of three RoboSpatial fine-tuned baselines: VILA~\citep{Lin_2024_CVPR}, SpaceLLaVA~\citep{chen2024spatialvlm} and LLaVA-NeXT~\citep{li2024llavanextstrong}, while remaining 1.59 percentage points below the 31.10\% reported for RoboPoint fine-tuned on RoboSpatial~\citep{song2025robospatial}. The RoboSpatial study fine-tunes these baselines on 900,000 spatial questions and 100,000 auxiliary grounding questions~\citep{song2025robospatial} whereas we use 40,000 spatial-context samples. The task coverage and data composition differ.

Our model also exceeds the reported performance of the foundation VLMs in our comparison, including GPT-5~\citep{singh2026openaigpt5card}, Gemini 2.5 Pro~\citep{comanici2025gemini25pushingfrontier} and Claude Sonnet 4.5~\citep{anthropic2025sonnet45}. Spatial-TRM scores 1.59 percentage points below the reported 31.10\% for Gemini Robotics-ER 1.5~\citep{geminiroboticsteam2025geminirobotics15pushing}, which is designed specifically for embodied reasoning and robotic spatial understanding. RoboBrain 2.0, SpaceTools, and RoboRefer achieve higher reported accuracies, ranging from 44.35\% to 61.48\% (\Cref{tab:robospatial-home}). 

Spatial-TRM contains only 11.2M trainable parameters and 877.8M in total, including its frozen encoders. Relative to RoboPoint's reported 13B size~\citep{yuan2025robopoint,song2025robospatial}, this is approximately $15\times$ fewer parameters. This reduction is in model size; inference speed and computational cost are not measured here.

\paragraph{Encoder choice controls the accuracy--size trade-off}
\begin{table}[h]
\caption{Encoder ablation on RoboSpatial-Home. CLIP halves the total parameter count relative to DINOv3, with a 3.28-percentage-point accuracy reduction. Parameter counts include the full model. Bold marks the best value in each numeric column.}
\label{tab:backbone-ablation}
\centering
\small
\setlength{\tabcolsep}{7pt}
\renewcommand{\arraystretch}{1.15}
\begin{tabular}{@{}lrrrr@{}}
\toprule
\textbf{Encoder} & \textbf{Accuracy (\%)} & \textbf{IoU} & \textbf{Dice} & \textbf{Parameters} \\
\midrule
DINOv3 ViT-L/16 & \textbf{29.51} & \textbf{0.11} & \textbf{0.16} & 877.8M \\
CLIP ViT-L/14 & 26.23 & 0.08 & 0.12 & \textbf{438M} \\
SigLIP-SO400M/14 & 16.39 & 0.07 & 0.12 & 888M \\
\bottomrule
\end{tabular}
\end{table}
We evaluate different encoder backbones while keeping the remaining architecture and training configuration unchanged. The CLIP ViT-L/14~\citep{radford2021clip} and SigLIP-SO400M/14~\citep{zhai2023siglip} variants achieve 26.23\% and 16.39\% context accuracy respectively, both with lower IoU and Dice scores indicating reduced spatial overlap with the ground truth, as shown in~\Cref{tab:backbone-ablation}. The CLIP variant reduces total parameter count from 877.8M to 438M (50.1\%) at a cost of 3.28 percentage points. SigLIP uses 888M parameters but scores 13.12 percentage points below DINOv3, showing that total size alone does not explain the differences between these encoders. This ablation also provides context for comparison with several baselines in~\Cref{tab:robospatial-home} that employ related vision backbones, including CLIP vision encoders in Molmo, LLaVA-Next, SpaceLLaVA, and RoboPoint, and SigLIP vision encoders in VILA and RoboRefer.

\paragraph{Sensitivity to evaluation criteria}
The context accuracy metric is designed for point-based predictions and requires selecting representative points from our predicted masks, making the score sensitive to that selection. The current public evaluator considers up to two points by default and accepts a prediction if any selected point is inside the target mask~\citep{song2026robospatialeval}; our headline result uses one predicted interior point. 

To quantify this effect, in addition to the one-point metric, we consider two-point, five-point, and ten-point criteria. The additional points are evenly sampled from the foreground pixels in row major order. A prediction is considered correct if any of the points lie within the ground truth mask. Context accuracy increases to 32.79\%, 37.70\% and 44.26\% respectively as more points are tested, a rise of 14.75 percentage points from one to ten points with the predicted masks unchanged. These increasingly permissive criteria measure sensitivity to point selection; the higher scores do not establish that the full predicted region is valid.

\paragraph{Across spatial relations}
We further break down context accuracy across the six spatial relationships. Spatial-TRM achieves 34.0\% for "front" (16/47), 30.0\% for "left" (6/20), 35.3\% for "right" (6/17), and 22.2\% for "behind" (2/9). 

Vertical relationships show the largest contrast: 35.7\% for "below" (5/14), compared with 6.7\% for "above" (1/15). The training data contains only 2.36\% "above" and "below" queries combined~(\Cref{app:training-data}), whereas they make up 23.8\% (29/122) of the benchmark questions. A relation-balanced training ablation would test whether this difference in coverage affects performance. 

Perspective ambiguity may also contribute to errors, particularly for objects with an intrinsic orientation whose reference frame may be ambiguous in the image.

\paragraph{Failure modes}
Small or distant objects are occasionally not grounded in cluttered scenes; in such cases, errors originate at the grounding stage, which prevents the model from anchoring a reference for subsequent spatial reasoning.

Also, while the predicted masks often capture the requested spatial relationship, they do not always respect the context of the scene and may intersect with other objects. The current objective does not explicitly supervise occupancy or collision avoidance.

%% file: text/08_discussion.tex
\section{Discussion}

Spatial-TRM separates pretrained feature extraction from task-specific spatial prediction. In the DINOv3 configuration, 98.7\% of the parameters belong to frozen encoders. Replacing them with CLIP halves the full model size for a 3.28-percentage-point accuracy reduction, making encoder choice a practical control over the model footprint. This model-size reduction does not establish a corresponding reduction in computation: the frozen encoders and repeated refinement steps still contribute to inference cost.

The output is a candidate region in the image. A single convex polygon imposes a clear geometric restriction: it cannot capture holes or disconnected valid regions. Nor does its shape establish that it is free of obstacles, provides physical support, or can accommodate a particular object. Using these predictions for robot actions would therefore require additional geometric and task-specific checks.

The supervision also limits what can be concluded from evaluation. RoboSpatial provides point annotations for spatial-context queries~\citep{song2025robospatial}; the derived masks need not describe the full set of valid locations. Point accuracy measures agreement at the selected coordinates, while mask overlap measures agreement with the extent of the derived target. Neither alone establishes the physical validity of an entire predicted region. Independent occupancy annotations and evaluation of more flexible region representations would help separate annotation mismatch from prediction error.

On RoboSpatial-Home, recursive refinement raises context accuracy from 20.49\% to 29.51\%: a gain of 9.02 percentage points (44.0\% relative) over a separately trained single-step model with the same 11.2M trainable parameters. The full system contains 877.8M parameters, approximately $15\times$ fewer than RoboPoint's reported 13B size~\citep{yuan2025robopoint,song2025robospatial}.

%% file: text/appendix.tex
\newpage
\appendix
\section{Appendix}

\subsection{Training Dataset}
\label{app:training-data}
We generate the RoboSpatial dataset using the generation pipeline described by~\citet{song2025robospatial}, restricting to indoor scenes from Matterport3D~\citep{Matterport3D}, 3RScan~\citep{Wald2019RIO} and ScanNetv2~\citep{dai2017scannet}. The original training data does not contain "above" and "below" relationships, although both are evaluated in RoboSpatial-Home. We extend the generation pipeline to include these relations: for "above", we sample points on the object’s top surface, while for "below", we sample points projected onto the surface beneath the object. 

We experiment with varying dataset sizes and find that performance improves with more samples, but the gains diminish when scaling beyond 40,000 samples. We therefore use 40,000 training samples and 999 validation samples, distributed approximately equally across the three datasets. All images are resized to $224 \times 224$ and normalized using ImageNet statistics. 

The resulting dataset contains 36,496 unique images in the training set and 995 in the validation set. 163 images from the training set also occur in the validation set paired with different anchor objects, relations and perspectives. To assess the effect of this overlap, we train an additional model after removing the 187 training samples corresponding to the shared images. With 39,813 training samples, the metrics on the unchanged validation set of 999 samples match those of the original run at the reported precision, with a Dice score of 0.24, IoU of 0.16, and centroid-in-GT accuracy of 28\%.

A total of 267 object classes are present in the training set, with a long tail distribution. The ten most frequent classes (door, window, cabinet, picture, table, shelf, chair, couch, doorframe, bed) account for 43.12\% of the training samples. The test set covers 143 of the same object classes. Spatial relations are distributed as: right (31.40\%), behind (26.23\%), left (22.19\%), front (17.82\%), below (1.46\%), above (0.90\%). The latter two are only present relative to the camera perspective. Camera, world, and object views account for 39.93\%, 38.03\%, and 22.03\% of the samples, respectively. The segmentation masks are highly imbalanced, with only 2.83\% of pixels belonging to the foreground class on average.

No additional filtering is applied to remove low-quality samples, and depth information is not utilized alongside RGB images, which could help resolve occlusion and depth ambiguity. Enhancing the diversity of the training dataset with tabletop scene data such as from HOPE~\citep{9981838} and GraspNet-1B~\citep{9156992} could further improve performance.

\subsection{Architecture Details}
\label{app:arch}
\subsubsection{Input Encoding} 
For an input image of size $224\times224$, the DINOv3 ViT-L/16 visual backbone produces a grid of $14\times14$ non-overlapping patches, together with one CLS token and four register tokens. For the downstream task, we discard these tokens and only retain the image patch features of shape $[B,196,1024]$. The dino.txt encoder tokenizes the text to a fixed context length of 77. We retain the second 1024-dimensional half aligned with the vision representation, resulting in features of shape $[B,77,1024]$ for both the anchor object and full-prompt embeddings.

\subsubsection{Feature Fusion}
Two separate learned linear layers map the image patch tokens and text tokens from 1024 dimensions to a common hidden dimension of 512. To fuse object-grounding information into the vision features, we apply a single multi-head cross-attention layer with the projected image features $[B,196,512]$ acting as queries and the projected anchor object text features $[B,x,512]$ as keys and values. Padding tokens in the text sequence are masked during attention. Learned query, key-value, and output projections are used, and scaled dot-product attention is computed with 8 heads of dimension 64. The attention output is added to the initial projected image features through a residual connection. These anchor-conditioned image features are concatenated along the sequence dimension with the projected full-prompt embeddings, forming an input latent $[B,273,512]$ for the subsequent reasoning module. No explicit grounding supervision is used, and the model relies on the pretrained dino.txt text-image alignment and downstream mask supervision to learn to ground the anchor object.

\subsubsection{Reasoning Block}
\label{app:reasoning} 
The reasoning module maintains two recurrent latent states, each with the same dimensionality as the input representation. The input latent is repeatedly injected into a reasoning state $z$, which is then used to update the current embedded answer $y$.

The backbone contains two transformer layers, each with a multi-head self-attention layer followed by a residual connection and RMS normalization~\citep{zhang2019rmsnorm}, and then a SwiGLU MLP~\citep{shazeer2020glu} followed by a second residual connection and RMS normalization. Self-attention uses eight heads with a dimension of 64. Every image and text token can attend to all 273 positions. The MLP has an expansion factor of four.

Rotary positional encodings~\citep{su2024roformer} are applied to the attention queries and keys, with a rotary base of 10,000. For the image patches, a two-dimensional rotary encoding represents their row and column coordinates in the $14\times14$ patch grid. The 77 text tokens use one-dimensional rotary encoding. 

Each latent state update executes the same two-layer transformer. During one deep recursion cycle, latent recursion updates the reasoning latent 4 times before one answer state update. There are 3 deep recursion cycles and only the final cycle retains gradients. Deep supervision repeats deep recursion 4 times with the option to exit early if the halting head predicts that the current answer state is sufficiently correct. The halting logit is trained with binary cross-entropy to predict whether the current mask satisfies the model's correctness criterion. A fixed exploration probability of 0.1 forces examples to continue until a minimum number of steps, randomly sampled between two and four. During evaluation, the halting head is disabled and the model executes the full deep-supervision budget.

\subsubsection{Decoder} 
The mask decoder projects the image tokens from the final embedded answer state and the initial DINOv3 image features to 256 dimensions, concatenates them, and projects back to 256 dimensions. The decoder context is combined with normalized patch coordinates, followed by RMS normalization. We use 32 half-planes as the upper bound for the number of polygon edges, although some may be redundant and may not form an edge. Each decoder layer first applies four-headed self-attention among the 33 learned queries for internal point and half-plane distances, followed by a residual connection and RMS normalization. It then applies cross-attention to allow the shape queries to attend to the decoder context, followed by another residual connection and RMS normalization. Finally, a SwiGLU MLP with an expansion factor of two is applied, followed by a third residual connection and RMS normalization. Two such decoder layers are stacked.

For example, $n_0 = (\cos 0, \sin 0) = (1, 0)$ defines the east boundary. For a pixel $p$,
\[
p_x \leq c_x + d_0
\]
where $c_x$ is the x-coordinate of the interior point and $d_0$ is the predicted offset from the interior point to the east boundary.

Both the interior point and half-plane distance heads are implemented as linear layers. The interior point undergoes a $\tanh$ activation to predict normalized coordinates constrained to the image bounds. The shared distance head uses a softplus activation to ensure positive offsets. The predicted convex polygon constructed using this half-plane boundaries determines the minimum margin for each pixel.

Since spatial coordinates are normalized to $[-1, 1]$, the minimum margin is converted to pixel units by multiplying by the reciprocal of the interval between adjacent pixel locations: $(S - 1) / 2$, where $S$ is the image size. To reduce sensitivity to boundary-localization errors, we scale the margins such that the predicted foreground probability transition smoothly from 0.1 to 0.9 over a distance of 10 pixels.

\subsubsection{Objective Functions}
\label{app:objectives}
We minimize the loss: $ 
\mathcal{L}=\mathcal{L}_{\mathrm{BCE}}
+0.5\mathcal{L}_{\mathrm{rel}}
+0.5\mathcal{L}_{\mathrm{halt}}
$ 
The binary cross-entropy positive-class weight is computed per batch as the ratio of negative to positive pixels. For the relational contrastive loss, the negative prompt is encoded with dino.txt and the margin between correct and negative relations is set to 0.25. Predictions are pooled to a $28\times28$ grid to assess spatial alignment, corresponding to an $8\times$ downsampling of the original $224\times224$ mask. The support points for the correctness criterion are computed at an offset of 3 pixels from the centroid. 

\subsection{Checkpoint Selection}
\label{app:checkpoint-selection}
We define a heuristic checkpoint selection score, using the validation centroid-in-GT accuracy adjusted for the generalization gap:
\[
S = 2 \times C_{\mathrm{val}} - C_{\mathrm{train}},
\]
where $C_{\mathrm{train}}$ and $C_{\mathrm{val}}$ denote the training and validation accuracies, respectively, measured as the percentage of predicted geometric foreground centroids that lie within the corresponding ground-truth masks. Benchmark context accuracy uses the decoder's predicted interior point instead of the geometric centroid. Across the five runs checked, the heuristic reproduces manual checkpoint selection based on increases in validation centroid-in-GT coinciding with decreases in validation loss. This criterion reflects the benchmark evaluation objective more directly than the loss formulation, which does not explicitly optimize whether the predicted centroid point lies inside the ground-truth region. The checkpoint with the highest adjusted validation score is selected for the final evaluation on RoboSpatial-Home.

\begin{figure}[t]
  \centering
  \includegraphics[width=\linewidth]{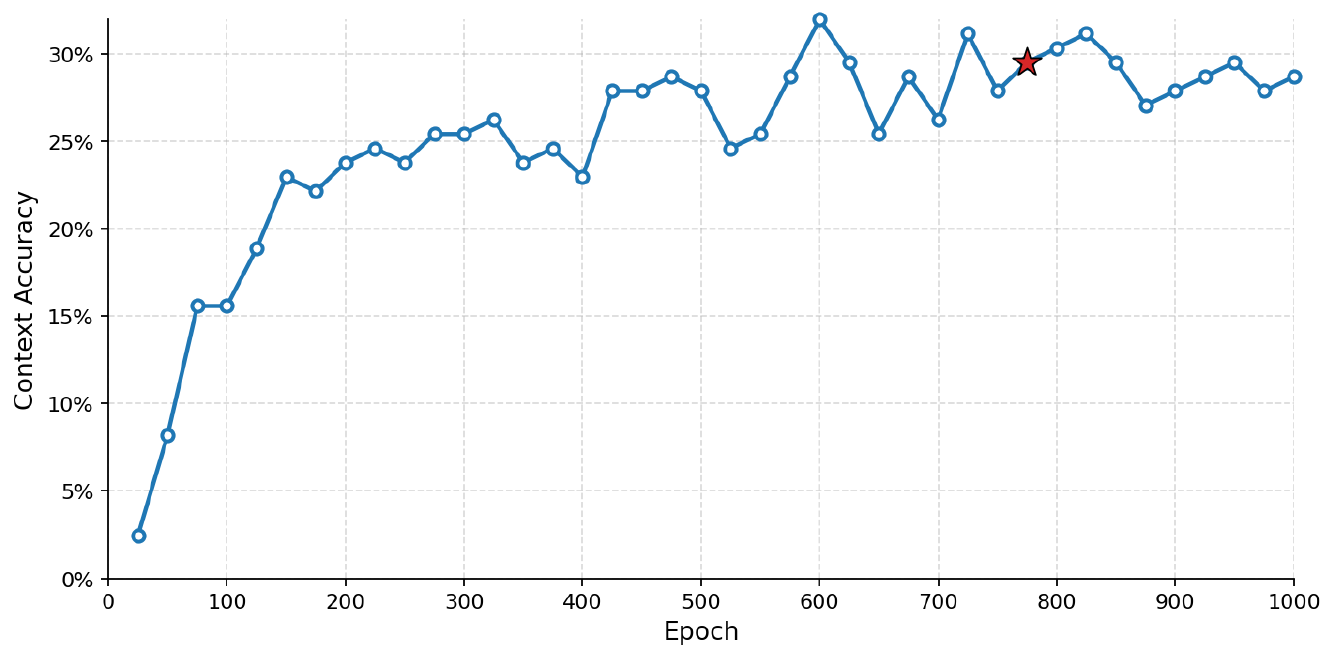}
  \caption{Context accuracy on the RoboSpatial-Home benchmark at 25-epoch intervals over the entire training duration. The checkpoint selected based on validation performance is marked with a red star.}
  \label{fig:benchmark-context-accuracy}
\end{figure}

We further evaluate context accuracy at checkpoints throughout training in~\Cref{fig:benchmark-context-accuracy}. RoboSpatial-Home contains only 122 context questions, making the fluctuations across checkpoints relatively coarse; each correctly answered question contributes approximately 0.82 percentage points to the overall accuracy. The validation-selected checkpoint scores 29.51\%, 2.46 percentage points below the 31.97\% peak at epoch 600, corresponding to a difference of three correctly answered questions. The benchmark peak is reported to illustrate checkpoint sensitivity and is not used to select the final checkpoint. 

\subsection{Training Setup}
\label{app:training-setup}
We train on a single node with four NVIDIA GH200 GPUs for 1000 epochs, with a batch size of 1536. We use Adam-atan2~\citep{everett2024scaling} with $\beta_1=0.9$, $\beta_2=0.95$ and weight decay 1.0. The learning rate linearly warms to $1e-4$ over 2000 steps and remains constant afterwards. We maintain an exponential moving average of parameters with decay 0.999 and evaluate every 25 epochs. Training uses bfloat16 precision and requires approximately 5 hours.

\section{Sensitivity to Recursion Budget}
We study the contribution of recurrence by varying the three dimensions of the recursion budget: (1) latent recursion steps, which update the reasoning latent, (2) deep recursion steps, which update the answer latent, and (3) deep supervision steps, which repeat deep recursion cycles. The checkpoint used for these post-training evaluations was trained with four latent recursion steps, three deep recursion steps and four deep supervision steps. In each ablation, only the specified dimension is varied, while the others are kept at their default values.

Executing the recursive model in non-recursive mode, with all three dimensions set to one, reduces context accuracy substantially from 29.51\% to 9.02\%. In contrast, reducing either deep recursion steps or deep supervision steps to one preserves performance at 30.33\% and 28.69\% respectively. However, setting both to one simultaneously lowers accuracy to 22.13\%, suggesting some redundancy across these two dimensions. Latent recursion is more sensitive, with a single latent recursion step reducing context accuracy to 14.75\%, indicating that refinement of the reasoning latent is particularly important for the model's performance. Conversely, we do not observe any gains from modestly increasing the recursion budget beyond the default values.

\section{Interpreting Region Evaluation}
\label{app:evaluation}
\begin{figure}[h]
  \centering
  \includegraphics[width=\linewidth]{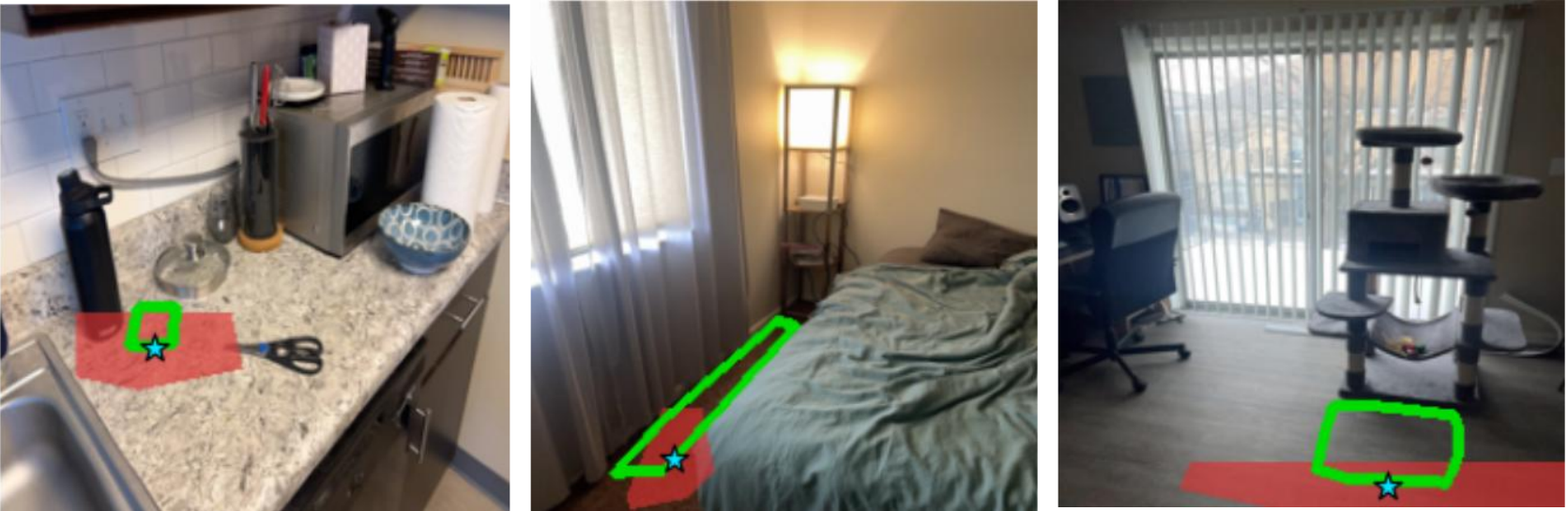}
  \caption{Predictions whose representative points fall outside the annotated target masks. Ground truth in green, predicted mask in red, and representative point marked with a star.}
  \label{fig:point-in-gt}
\end{figure}

\begin{figure}[h]
  \centering
  \includegraphics[width=\linewidth]{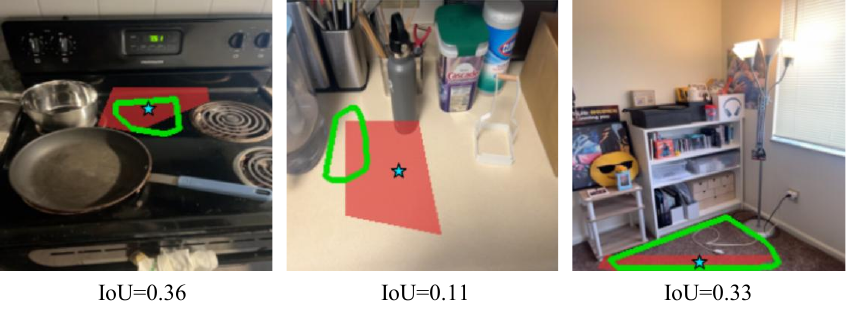}
  \caption{Examples of predicted regions with limited overlap with the annotation-derived targets. Ground truth masks shown in green, predicted masks shown in red.}
    \label{fig:mask-overlap}
\end{figure}

\Cref{fig:point-in-gt,fig:mask-overlap} illustrate how representative-point selection and annotation extent affect the reported metrics. These examples do not establish that the entire predicted region is vacant or spatially valid.

\subsection{Implications for Training}
The under-constrained ground truth masks also introduce a challenge during optimization. Standard segmentation losses penalize deviations from the geometry of the annotations. Overlap-based losses directly encourage agreement with the annotated region. Binary cross-entropy also supervises agreement with the annotations at each pixel, although it does not explicitly optimize for global overlap.

Conversely, an objective that only penalizes the direction and perspective of the predicted region fails to supervise the extent and context of vacant space.

A future objective could supervise direction and vacancy while allowing multiple valid regions. Developing such an objective remains a possible direction for future research.